\pdfoutput=1

\documentclass[11pt]{article}

\usepackage[]{acl}
\usepackage{booktabs}
\usepackage{multirow}
\usepackage{amssymb}
\usepackage{times}
\usepackage{amsmath}
\usepackage{latexsym}
\usepackage{graphicx}
\usepackage{algorithmic}
\usepackage{algorithm}
\usepackage{subfigure}
\usepackage{xcolor}
\usepackage{pifont}

\usepackage[T1]{fontenc}

\usepackage[utf8]{inputenc}

\usepackage{microtype}

\usepackage{inconsolata}

\title{
LLatrieval: LLM-Verified Retrieval for Verifiable Generation
}

\author{
Xiaonan Li\thanks{\ \ Equal Contribution}\ , Changtai Zhu$^*$, Linyang Li, Zhangyue Yin, Tianxiang Sun, Xipeng Qiu \\
School of Computer Science, Fudan University\\
Shanghai Key Laboratory of Intelligent Information Processing, Fudan University \\
\{lixn20, xpqiu\}@fudan.edu.cn, \ ctzhu23@m.fudan.edu.cn
} 

\begin{document}
\maketitle
\begin{abstract}
Verifiable generation aims to let the large language model (LLM) generate text with supporting documents, which enables the user to flexibly verify the answer and makes the LLM's output more reliable. Retrieval plays a crucial role in verifiable generation. Specifically, the retrieved documents not only supplement knowledge to help the LLM generate correct answers, but also serve as supporting evidence for the user to verify the LLM's output.
However, the widely used retrievers become the bottleneck of the entire pipeline and limit the overall performance. Their capabilities are usually inferior to LLMs since they often have much fewer parameters than the large language model and have not been demonstrated to scale well to the size of LLMs. 
If the retriever does not correctly find the supporting documents, the LLM can not generate the correct and verifiable answer, which overshadows the LLM's remarkable abilities.
To address these limitations, we propose \textbf{LLatrieval} (\textbf{L}arge \textbf{La}nguage Model Verified Re\textbf{trieval}),
where the LLM updates the retrieval result until it verifies that the retrieved documents can sufficiently support answering the question. 
Thus, the LLM can iteratively provide feedback to retrieval and facilitate the retrieval result to fully support verifiable generation.
Experiments on ALCE show that LLatrieval significantly outperforms extensive baselines and achieves state-of-the-art results. 
\end{abstract}

\section{Introduction}
\begin{figure}[t]
    \centering
    \includegraphics[width=0.4\textwidth]{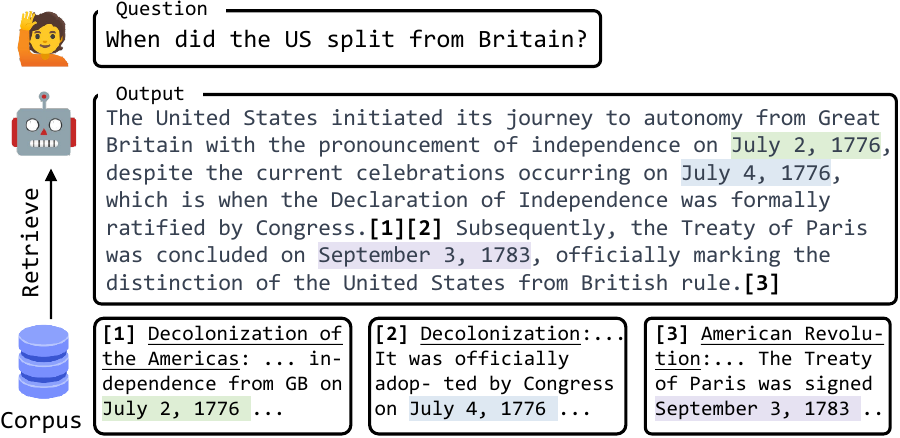}
    \caption{Verifiable Generation~\citep{alce}}
    \label{fig:verifiable_qa}
\end{figure}

Large language models (LLMs) have shown remarkable abilities over various downstream tasks~\citep{gpt4_openai,instruct_gpt,palm,llama2,llm_survey}. However, LLMs struggle with factual errors, and often produce non-factual and fabricated content~\citep{hallucination_survey_tencent, hallucination_survey_foundation_model,factuality_survey,hlt_hallucination_survey,fact_score,factool}, which is usually referred  to  as ``hallucination'' and makes the LLM's response not trustworthy. 
To address these challenges, researchers propose a new generation paradigm, Verifiable Generation~\citep{rarr,attributed_qa_for_llm,alce,attribution_survey}, shown in Figure~\ref{fig:verifiable_qa}. For a given question, it requires the LLM to generate the answer with corresponding supporting documents.
In this way, these documents can serve as evidence and enable users to flexibly verify the answer, which makes the LLM's response more reliable and facilitates its application in various important scenarios, medical diagnosis~\citep{verifiable_generation_medical_diagnosis}, scientific writing~\citep{verifiable_generation_scientific_writing}, situation reports~\citep{verifiable_generation_situation_reports}, etc.

\begin{figure}[t]
    \centering
    \subfigure[Vanilla Retrieval]{
        \includegraphics[width=0.3\textwidth]{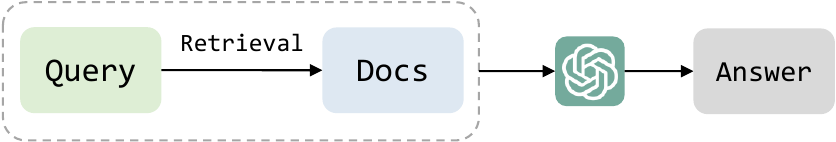}
        \label{fig:vannila_retrieval_for_verifiable_generation}
    }
    \subfigure[LLatrieval]{
        \includegraphics[width=0.4\textwidth]{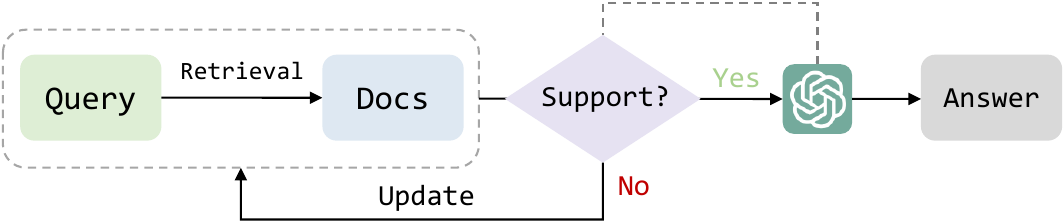}
        \label{fig:our_method_overview}
    }
    \caption{When the vanilla retrieval overshadows the LLM's remarkable abilities in the pipeline, LLatrieval can fully harness the LLM's abilities to the retrieval by verify-update iterations.}
\end{figure}
Retrieval plays a crucial role in verifiable generation. Take retrieval-read, a typical pipeline for verifiable generation~\citep{alce,attributed_qa_for_llm}, as an example, shown in Figure~\ref{fig:vannila_retrieval_for_verifiable_generation}.  It is divided into two stages:  1) retrieve the relevant documents for the given question, 2) generate the answer according to the retrieved documents.
On the one hand, the retrieved documents serve as knowledge supplement to help the LLM generate the answer. On the other hand, the answer's verifiability relies on the retrieved documents.
Thus, both of the answer's correctness and verifiability highly rely on the quality of retrieval result.

However, the widely used retrievers become the bottleneck of the entire pipeline and limit the overall performance~\citep{context_filtering_for_rag,making_rag_robust_to_irrelevant_context}. 
Specifically, although the dense retrievers have achieved a series of state-of-the-art results in various scenarios~\citep{dpr,ar2,instructor}, their capabilities are substantially inferior to LLMs~\citep{gpt3,palm},
since they usually have much fewer parameters, and have not been demonstrated to scale well to the size of LLMs~\citep{gpt3} and achieve comparable capabilities.
If the retriever does not correctly find the supporting documents, it is challenging for the LLM to output the answer which is both correct and verifiable. 
In the typical pipeline, the LLM usually receives the retrieval result in a passive manner and can not provide feedback to the low-quality retrieval even if it is capable of identifying that the retrieved documents are irrelevant to the question. These make the retrieval overshadow the LLM's remarkable abilities and limit the overall performance.

To address these limitations, we propose the framework of \textbf{LLatrieval} (\textbf{L}arge \textbf{La}nguage Model Verified Re\textbf{trieval}), shown in Figure~\ref{fig:our_method_overview}, where the LLM fully provides feedback to the retrieval and augments it by the verify-update iteration: 1) Retrieval Verification: The LLM verifies whether the retrieval result can support answering the give question. For questions whose documents fail the verification, we will update their documents and thus avoid low-quality retrieval.
2) Retrieval Update: 
The LLM updates and improves the retrieval result. Specifically, we propose Progressive Selection and Missing-Info Query to let the LLM progressively scan document candidates and supplement missing information to facilitate better verifiable generation.
Through the verify-update iteration, the LLM can iteratively refine the retrieval result until it verifies that the retrieved documents can support answering the given question, and thus generate both correct and verifiable answers.

We summarize our contributions as follows:
\begin{itemize}
    \item To the best of our knowledge, LLatrieval is the first framework to let the LLM fully provide feedback to retrieval and thus augments it through verify-update iterations.
    \item We conduct comparison with extensive baselines and results show that LLatrieval significantly outperforms baselines and achieves new state-of-the-art results on verifiable generation. 
    \item Further analyses show that each component contributes critically to improvements and our method shows the potential to scale retrieval by scaling the LLM.
    \item We make our code publicly available to facilitate future research.
\footnote{\href{https://github.com/BeastyZ/LLM-Verified-Retrieval}{https://github.com/BeastyZ/LLM-Verified-Retrieval}}
\end{itemize}

We hope that LLatrieval can inspire researchers of the important design choices for LLM-augmented Retrieval and pave the way for further improvements.

\section{\mbox{Background: Verifiable Generation}}
In this paper, we follow the setting proposed by \citet{alce}, which requires the LLM to generate the answer with citations to the supporting documents, as shown in Figure~\ref{fig:verifiable_qa}. This setting is close to the real-world applications of generative search engines, e.g., New Bing, and the citations allow the user to verify the answer more conveniently.
We introduce its typical pipeline and the verifiability evaluation as follows. 
\subsection{Pipeline}
Verifiable generation is usually demonstrated by retrieval-read~\citep{alce,attributed_qa_for_llm,expert_qa}, divided into two stages: retrieval and generation. 
First, given the question $q$, the relevant documents, $D=\{d_1, d_2,\cdots, d_k\}$, are retrieved as follows:

\begin{small}
\begin{align}
    D=\operatorname{R}(q,\mathbb{D},k) =\operatorname{Top-\textit{k}}_{d\in \mathbb{D}} \operatorname{score}(q,d)
\end{align}
\end{small}
where $\operatorname{R}$ is the retriever and 
$\operatorname{score}(\cdot,\cdot)$ is the similarity score of $\operatorname{R}$~\citep{bm25,dpr}. 
Then the LLM leverages these documents to generate the answer as:

\begin{small}
\begin{align}
    A = \operatorname{LLM}(I_g;Demos;q;D),
    \label{eq:llm_answer_generation}
\end{align}
\end{small}
where we follow \citet{alce} to use the instruction, $I_g$, and demonstrations, $Demos$, to make the LLM generate answer and citations to $D$. 

\subsection{Verifiability Evaluation}
\label{sec:citation_eval}
Following \citet{alce,evaluating_verifiability_generative_search_engine}, we evaluate the answer's verifiability by measuring citation quality: \textbf{Citation Recall} evaluates whether the output is fully supported by cited documents; \textbf{Citation Precision} identifies those citations which can not help support the answer. According to them, we calculate \textbf{Citation F1} to comprehensively evaluate the answer's verifiability. 
Since the retrieved documents determine the citation quality and the answer's verifiability, the retrieval is crucial for the overall pipeline.
We adopt the recently proposed ALCE benchmark for automatic citation evaluation~\citep{alce}.

\section{LLatrieval: LLM-Verified Retrieval}
In the retrieval-read pipeline, the LLM passively receives the retrieval result and can not provide feedback to the low-quality retrieval, which makes the large language model with remarkable abilities overshadowed by the retriever with much fewer parameters.
To address these limitations, we propose LLatrieval, where the LLM iteratively provides feedback to the retrieval through verify-update iterations, and thus fully harnesses its abilities for verifiable generation. In this section, we introduce the modules of retrieval verification and update, and the iterative verify-update process as follows.

\subsection{Retrieval Verification}
\begin{figure}[t]
    \centering
    \includegraphics[width=0.4\textwidth]{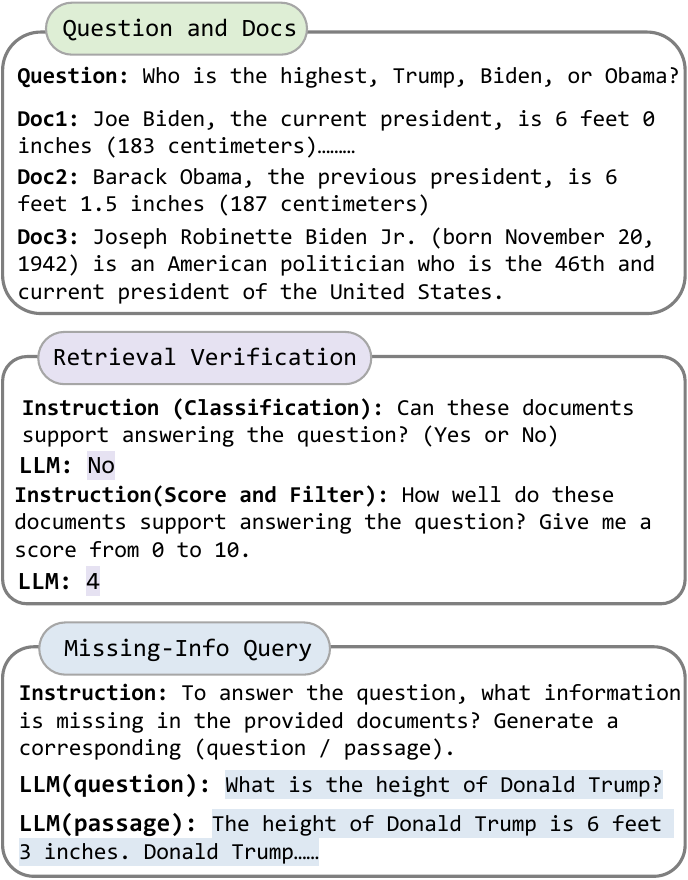}
    \caption{\textbf{Retrieval Verification:} We propose two ways to verify whether the documents can support answer the question. \textbf{Missing-Info Query:} We propose to let the LLM generate missing-info query in two styles.
    }
    \label{fig:retrieval_verification_and_miss_info_query}
\end{figure}
Since low-quality retrieval overshadows the LLM's remarkable abilities for verifiable generation, we propose to use the LLM itself to identify the low-quality retrieval and thus avoid it. 
In this paper, we instantiate the retrieval verification as two simple instruction-based methods, shown in Figure~\ref{fig:retrieval_verification_and_miss_info_query}, to let the LLM verify whether the retrieved documents can support answering the question, and leave more elaborate methods as future work.
\paragraph{Classification}
Given the question $q$ and retrieved documents $D$, we verify whether $D$ can sufficiently support answering $q$ by prompting LLM to give a binary label (\textit{yes} or \textit{no}) as:

\begin{small}
\begin{align}
    \text{Verify-Result} = \operatorname{LLM}(I_v^c;q;D)
\end{align}
\end{small}
where $I_v^c$ is the corresponding instruction.
Through the LLM's strong natural language understanding ability, we can fully identify the low-quality retrieval and thus avoid it. 
Meanwhile, LLatrieval finishes when the example's retrieved documents pass the retrieval verification, which enables it to dynamically adjust the LLM cost under varying retrieval demands.
\paragraph{Score-and-Filter} We further explore a score-and-filter strategy for retrieval verification. We first score the current documents by how well they can support answering the question and then filter low-quality retrieval by the pre-defined threshold $\tau$ as:

\begin{small}
\begin{align}
    \text{Score}_D &= \operatorname{LLM}(I_v^s;q;D)\\
    \text{Verify-Result} &= \mathrm{Yes} \ \operatorname{if} \ \text{Score}_D \geq \tau \ \ \operatorname{else} \ \mathrm{No}
\end{align}
\end{small}
where $I_c^s$ is the corresponding instruction. The higher $\tau$ leads to stricter retrieval verification criterion, better quality of the final retrieval result and more iterations. By tuning $\tau$, we can dynamically adjust the strictness of verification and the corresponding LLM cost.

\subsection{Retrieval Update}

When the current documents fail the retrieval verification, we will update and refine the low-quality retrieval result.
Specifically, we propose the following two LLM-based modules to update the retrieval result in synergy and thus better harness the LLM's abilities for retrieval.

\paragraph{Progressive Selection}
\begin{figure}[t]
    \centering
        \subfigure[Progressively selecting the documents, where the blue and purple color represent the current documents $D$ and new candidates $D_c^*$ in each selection.]{
    \includegraphics[width=0.4\textwidth]{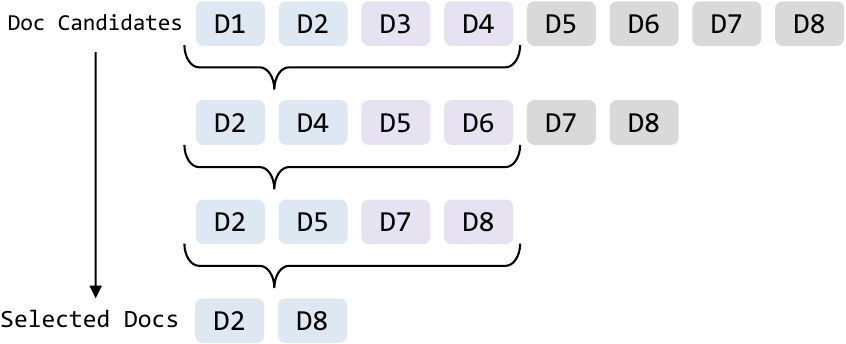}
    \label{fig:progressive_selection_sliding_window}
    }
    \subfigure[Using the LLM to select docs from candidates.]{
    \includegraphics[width=0.4\textwidth]{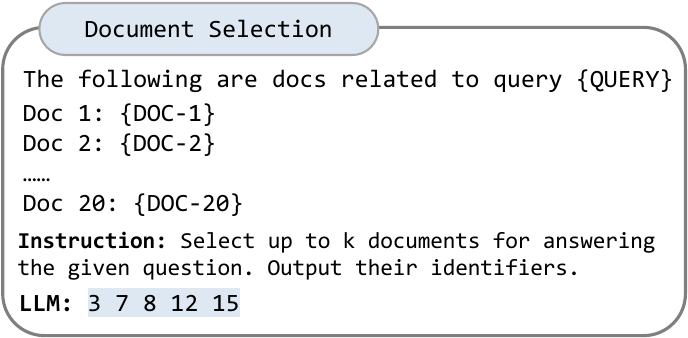}
    \label{fig:progressive_selection_llm_select}
    }
    \caption{
    Retrieval update by progressive selection.
    }
    \label{fig:progressive_selection}
\end{figure}

Since the dense retrievers~\citep{dpr,instructor} measure similarity based on the dual-encoder architecture and usually have fewer parameters than the prevailing large language models~\citep{gpt3, palm}, it is challenging for them to fully demonstrate fine-grained token-level interactions and thus accurately rank the most relevant documents at top-K~\citep{ar2}. Inspired by the retrieval-rerank framework~\citep{mono_bert}, which reranks the document candidates returned by the retriever to get the final retrieval result, 
we propose Progressive Selection for retrieval update: the LLM progressively scans the document candidates returned by the retriever and selects the supporting documents, shown in Figure~\ref{fig:progressive_selection}.
For each selection, we obtain a small list of new document candidates, $D_c^*$, by sliding window on the document candidates (Figure~\ref{fig:progressive_selection_sliding_window}). Then we use the LLM to select $k$ documents which can maximally support answering $q$, from $D \cup D_c^*$, as the \textit{updated} $D$ (Figure~\ref{fig:progressive_selection_llm_select}). 
Thus, the LLM can remove irrelevant documents in $D$ and integrate the critical information in $D_c^*$ to update and improve $D$, with $D$'s size unchanged. Compared to reranking, the selection directly outputs a combination of documents,
so the updated $D$ can avoid documents with redundant information and thus get more comprehensive.
With iteratively updating $D$ from new $D_c^*$, the LLM can progressively improve $D$ to make it better suppport verifiable generation.

\paragraph{Missing-Info Query}
Since the current documents can not sufficiently support answering the question, we propose to update the current documents $D$ by querying missing information of $D$ for supplement, shown in Figure~\ref{fig:retrieval_verification_and_miss_info_query}. Specifically, we first use the LLM to generate the query of the missing information in $D$ for answering $q$ as:

\begin{small}
\begin{align}
    Q = \operatorname{LLM}(I_{q}; q; D),
\end{align}
\end{small}
where $I_q$ instructs the LLM to identify the missing information in the current documents $D$ for the given question $q$ and generate the corresponding query $Q$, shown in Figure~\ref{fig:retrieval_verification_and_miss_info_query}. Then we use $Q$ to retrieve the corresponding documents for information supplement. For the fusion of the newly retrieved documents and $D$, please refer to section~\ref{sec:update_verify_iteration}.

When the current dense retrievers often use the BERT-based architecture~\citep{dpr} and produce one vector for calculating similarity, it is challenging for them to comprehensively tackle the user query involving the knowledge from multiple aspects~\citep{sub_sentence_encoder,multi_view_dense_encoder} and complicated reasoning, which widely exists in the real-world scenarios. 
During the verify-update iterations, the missing-info query can supplement multiple-aspect information through iteratively identifying and querying the missing information by LLM, which can fully harness the LLM's remarkable language understanding and reasoning abilities. 
In experiments, we explore two styles of missing-info query: 1) \textit{question}: the corresponding question of missing information; 2) \textit{pseudo passage}: inspired by the query rewriting methods~\citep{hyde,lamer}, we generate a pseudo passage of missing information as retrieval query, which neighbors the relevant documents and thus can help better retrieve them. We regard developing stronger missing-info query modules as future work.

\begin{algorithm}[t]
\small
	\begin{algorithmic}[1]
 \renewcommand{\COMMENT}[2][.1\linewidth]{\leavevmode\makebox[#1][l]{\#~#2}}
 \caption{LLatrieval}
 \label{alg:llm_verified_retrieval}
 \label{alg_progressive_filtering}
  	\renewcommand{\algorithmicrequire}{\textbf{Input:}}
	\renewcommand{\algorithmicensure}{\textbf{Output:}}
 
  	\REQUIRE Question $q$, document pool $\mathbb{D}$, the large language model $\operatorname{LLM}$, the retriever $\operatorname{R}$, the maximum iteration $T$, each iteration's document candidates quantity $N$
   \ENSURE Supporting Documents $D$
    \STATE $Q\gets q$
    \STATE $D\gets \{\}$
    \FOR{$i$ in $1,2\cdots T$}

    \IF{$D\ne\{\}$}
    \STATE 
$Q\gets$ Use LLM to generate missing-info's query
    \ENDIF

    \STATE $D_c\gets R(Q,\mathbb{D}, N)$
    \FOR{$D_c^*$ in SlidingWindow($D_c$)}
    \STATE{$D\gets$ Use LLM to select $k$ docs from $D \cup D_c^*$}
    \ENDFOR

    \IF{ $\operatorname{Verify}(q,D) \to Yes$ }
    \STATE \COMMENT{$D$ can support answering $q$, so return $D$.}
    \STATE break
    \ENDIF
    
    \ENDFOR
    \RETURN $D$

	\end{algorithmic} 
\end{algorithm}

\subsection{Verify-Update Iteration}
\label{sec:update_verify_iteration}
We show the overall procedure in Algorithm~\ref{alg:llm_verified_retrieval}.
In each iteration, we first update the missing-info query and retrieve a new list of document candidates $D_c$ that may contain missing information of $D$ (Line $4 \sim 7$). Then we update the current $D$ by progressively selecting from $D_c$ (Line $8 \sim 10$). 
Through the synergy between progressive selection and missing-info query, we can identify and retrieve $D$'s missing information and progressively incorporate them into $D$.
Since the progressive selection is conducted on $D \cup D_c$, we can not only supplement $D$'s missing information for $q$, but also fully retain the critical documents in $D$. In this way, we comprehensively integrate the relevant documents with multifaceted information and thus support better verifiable generation.
After each update, we use the LLM to verify whether the current documents $D$ can support answering the given question $q$ (Line $11 \sim 13$) and continue updating the low-quality retrieval result in the next iteration.
Through the verify-update iteration, the retrieval result can be iteratively updated and improved until it can sufficiently support high-quality verifiable generation, and thus help the LLM generate both correct and verifiable answers.
Meanwhile, our method can dynamically adjust the number of iterations under varying retrieval demands and thus help save the LLM cost.

\section{Experiments}
\subsection{Experimental Settings}
\label{sec:experimental_settings}

\begin{table*}[t]
\centering
\setlength\tabcolsep{4pt}
\small
\begin{tabular}{@{}lcccccccccccccc@{}}
\toprule
\textbf{Dataset}                                                                      & \multicolumn{4}{c}{\textbf{ASQA}}                                                         & \multicolumn{4}{c}{\textbf{QAMPARI}}                                                      & \multicolumn{4}{c}{\textbf{ELI5}}                                                         & \multicolumn{2}{c}{\textbf{Overall}}                                                                                \\ \midrule
\multirow{2}{*}{\textbf{\begin{tabular}[c]{@{}l@{}}Evaluation\\ Metric\end{tabular}}} & \textbf{\hspace{-5pt}Correct\hspace{-5pt}}     & \multicolumn{3}{c}{\textbf{Citation}}                              & \textbf{\hspace{-5pt}Correct\hspace{-5pt}}     & \multicolumn{3}{c}{\textbf{Citation}}                              & \textbf{\hspace{-5pt}Correct\hspace{-5pt}}     & \multicolumn{3}{c}{\textbf{Citation}}                              & \multirow{2}{*}{\textbf{\hspace{-5pt}Correct\hspace{-5pt}}} & \multirow{2}{*}{\textbf{\begin{tabular}[c]{@{}c@{}}Citation\\ F1\end{tabular}}} \\ \cmidrule(lr){2-13}
& \textbf{EM-R}        & \textbf{Rec}         & \textbf{Prec}        & \textbf{F1}          & \textbf{F1}          & \textbf{Rec}         & \textbf{Prec}        & \textbf{F1}          & \textbf{Claim}       & \textbf{Rec}         & \textbf{Prec}        & \textbf{F1}          &                                   &                                                                                 \\ \midrule
BM25                                                                                  & 51.7                 & 43.3                 & 48.5                 & 45.7                 & 17.5                 & 21.4                 & 25.7                 & 23.3                 & 11.7                 & 52.9                 & 49.0                 & 50.9                 & 27.0                              & 40.0                                                                            \\
DPR                                                                                   & 54.4                 & 52.2                 & 54.6                 & 53.3                 & 18.5                 & 20.0                 & 23.2                 & 21.5                 & -                    & -                    & -                    & -                    & -                                 & -                                                                               \\
Contriever-MS                                                                         & 54.9                 & 54.0                 & 57.1                 & 55.5                 & 17.8                 & 20.7                 & 23.7                 & 22.1                 & -                    & -                    & -                    & -                    & -                                 & -                                                                               \\
GTR                                                                                   & 54.4                 & 52.2                 & 54.6                 & 53.3                 & 18.5                 & 20.0                 & 23.2                 & 21.5                 & -                    & -                    & -                    & -                    & -                                 & -                                                                               \\
Instructor-large                                                                      & 55.6                 & 52.1                 & 54.1                 & 53.1                 & 17.2                 & 19.4                 & 21.9                 & 20.6                 & -                    & -                    & -                    & -                    & -                                 & -                                                                               \\
BGE-E-large                                                                           & 55.9                 & 56.4                 & 58.6                 & 57.5                 & 17.3                 & 22.7                 & 25.5                 & 24.0                 & -                    & -                    & -                    & -                    & -                                 & -                                                                               \\ \midrule
monoBERT (50)                                                                         & 53.8                 & 55.8                 & 57.4                 & 56.6                 & 20.9                 & 26.6                 & 30.3                 & 28.3                 & 13.8                 & 63.5                 & 58.3                 & 60.8                 & 29.5                              & 48.5                                                                            \\
monoBERT (100)                                                                        & 53.0                 & 54.0                 & 56.5                 & 55.2                 & 20.2                 & 25.5                 & 29.1                 & 27.2                 & 14.1                 & 63.3                 & 58.2                 & 60.6                 & 29.1                              & 47.7                                                                            \\
monoT5-3B (50)                                                                        & 54.0                 & 55.6                 & 57.7                 & 56.6                 & 20.3                 & 28.1                 & 32.0                 & 30.0                 & 13.9                 & 64.2                 & 59.3                 & 61.7                 & 29.4                              & 49.4                                                                            \\
monoT5-3B (100)                                                                       & 53.7                 & 53.0                 & 55.2                 & 54.1                 & 19.7                 & 26.0                 & 29.6                 & 27.7                 & 14.1                 & 63.9                 & 58.8                 & 61.2                 & 29.2                              & 47.7                                                                            \\
BGE-R-large (50)                                                                      & 55.3                 & 57.2                 & 59.0                 & 58.1                 & 20.7                 & 26.6                 & 30.8                 & 28.5                 & 13.9                 & 65.9                 & 61.4                 & 63.6                 & 30.0                              & 50.1                                                                            \\
BGE-R-large (100)                                                                     & 54.7                 & 57.4                 & 58.7                 & 58.1                 & 19.9                 & 25.9                 & 29.6                 & 27.6                 & 14.4                 & 67.1                 & 62.6                 & 64.8                 & 29.7                              & 50.2                                                                            \\ \midrule
HYDE                                                                                  & 56.8                 & 59.1                 & 60.5                 & 59.8                 & 16.2                 & 20.2                 & 27.7                 & 23.4                 & 15.2                 & 70.4                 & 63.4                 & 66.7                 & 29.4                              & 49.9                                                                            \\
HYDE*                                                                                 & 56.7                 & 57.5                 & 57.4                 & 57.5                 & 16.5                 & 20.1                 & 26.2                 & 22.7                 & -                    & -                    & -                    & -                    & -                                 & -                                                                               \\
LAMER                                                                                 & 55.3                 & 59.1                 & 60.1                 & 59.6                 & 19.0                 & 23.0                 & 31.0                 & 26.4                 & 14.4                 & 70.8                 & 62.7                 & 66.5                 & 29.0                              & 50.0                                                                            \\
LAMER*                                                                                 & 55.7                 & 58.7                 & 60.2                 & 59.4                 & 16.8                 & 20.6                 & 29.1                 & 24.1                 & -                    & -                    & -                    & -                    & -                                 & -                                                                               \\
RankGPT (50)                                                                          & 56.4                 & 58.5                 & 58.9                 & 58.7                 & 20.7                 & 27.8                 & 31.6                 & 29.6                 & 13.5                 & 64.7                 & 58.8                 & 61.6                 & 30.2                              & 50.0                                                                            \\
RankGPT (100)                                                                         & 57.0                 & 58.8                 & 59.7                 & 59.2                 & 22.0                 & 28.4                 & 32.9                 & 30.5                 & 13.5                 & 63.5                 & 57.6                 & 60.4                 & 30.8                              & 50.1                                                                            \\ \midrule
LLatrieval (passage) \hspace{-20pt}                                                                  & \textbf{57.7}        & 60.9                 & 61.3                 & \textbf{61.1}        & \textbf{24.3}        & 32.4                 & 36.6                 & \textbf{34.4}        & 16.5                 & 75.8                 & 67.4                 & 71.3                 & \textbf{32.9}                     & \textbf{55.6}                                                                   \\
LLatrieval (question) \hspace{-20pt}                                                                & 57.3                 & 60.5                 & 60.6                 & 60.6                 & 23.6                 & 30.9                 & 34.8                 & 32.7                 & \textbf{16.7}        & 75.4                 & 68.0                 & \textbf{71.5}        & 32.5                              & 54.9                                                                            \\ \bottomrule
\end{tabular}
\caption{The performance comparison on ALCE~\citep{alce}, where ``(passage)'' and ``(question)'' represent the styles of missing-info query in LLatrieval. For reranking baselines, the number in parentheses is the number of reranked document candidates.
For query rewriting baselines, the superscript ``*'' represnets using their original retriever, otherwise using ours.
}
\label{tab:main_result}
\end{table*}
\paragraph{Method Comparison}

Since we focus on augmenting the retrieval for verifiable generation, we compare LLatrieval with various retrieval methods by evaluating the LLM output's correctness and verifiability under the same retrieval-read pipeline~\citep{alce}, including:
\textbf{Retriever}: 1) BM25~\citep{bm25} 2) DPR~\citep{dpr} 3) Contriever~\citep{contriever}: A large-scale unsupervised contrastive learned dense encoder~\citep{contriever}. 
4) GTR~\citep{gtr_embedding}:
A large dense retriever initialized from T5-XXL, pre-trained on a web-mined corpus and fine-tuned on information retrieval datasets. 
5) Instructor~\citep{instructor}: An instruction-tuned dense retriever on about 330 diverse tasks. 
6) BGE-Embedding~\citep{bge_embedding}: A recent large-scale trained general-purpose embedder which outperforms OpenAI's ``text-embedding-ada-002'' API.
\textbf{Re-ranker}: 1) monoBERT~\citep{mono_bert}\&monoT5~\citep{mono_t5}: traditional point-wise re-rankers based on BERT and T5. 
2) BGE-Reranker~\citep{bge_embedding}: A recent large-scale trained general purpose re-ranker.
\textbf{LLM-based Retrieval Augmentation}:
1) Query Rewriting: We follow HYDE~\citep{hyde} and LAMER~\citep{lamer} to generate the pseudo passage based on the question and its relevant passages, respectively. Then we use the pseudo passage to conduct retrieval. 
2) LLM-reranking: RankGPT~\citep{rank_gpt}, which prompts the LLM to rerank passages by permutation generation. 
Except the retriever baselines, we use the same retriever as in LLatrieval. For more details, please refter Appendix~\ref{appendix:implementation_details}. We show the correlation between retrieval quality and generation quality in Appendix~\ref{appendix_correlation_retrieval_generation}.

\paragraph{Dataset and Evaluation}
We conduct experiments on the ALCE benchmark~\citep{alce}, which consists of three long-form QA datasets, and follow ALCE to evaluate the correctness and verifiability of generation.

\textbf{ASQA}~\citep{asqa} is a long-form factoid QA dataset where each question requires multiple short answers to cover the multiple aspects of it. To evaluate the answer's correctness, we calculate the recall of correct short answers (from the dataset) by identifying whether the short answers match substrings of model output (\textit{exact match recall}; EM-R).

\textbf{QAMPARI}~\citep{qampari}: a factoid QA dataset where the answer is a list of entities from different documents. To evaluate the answer's correctness, we calculate the 
F1
of the prediction by identifying the exact match to the gold answer list. Following ~\citet{alce}, we consider the recall to be 100\% when prediction covers at least 5 correct answers.

\textbf{ELI5}~\citep{eli5}: a long-form QA dataset where each question requires comprehensive long answers and multiple documents as evidence. To evaluate the answer's correctness, we follow \citet{alce} to measure whether the model prediction entails the sub-claims of the gold answer.

Following ALCE~\citep{alce}, on ASQA and QAMPARI, we use aliases of short answers provided by the dataset and normalize the model output and the short answers when measuring exact match.
For ASQA, we use its sub-questions as the question to eliminate the original question's ambiguity, for simplicity. 

For detailes of verifiabiliy evaluation, please refer to Section~\ref{sec:citation_eval}.

\paragraph{Implementation Details}
We use the public OpenAI API of ``gpt-3.5-turbo-0301'' for retrieval and generation, unless otherwise specified. 
We use window size=20 and set the number of document candidates per query as 50. 
We set the number of supporting documents and the maximum of iterations as 5 and 4. 
We show LLatrieval's performance under different hyper-parameters and retrievers in Appendix~\ref{appendix_method_sensitivity}. For the document pool, we follow \citet{alce} to use the Wikipedia for ASQA and QAMPARI, and Sphere~\citep{spere}, a filtered version of a filtered version of Common Crawl
for ELI5, respectively.
For ASQA and QAMPARI, we use the recently proposed dense embedder, BGE-large~\citep{bge_embedding}, as the retriever. For ELI5, we follow \citet{alce} to use BM25~\citep{bm25} since dense retrievers are costly and slow for large-scale web corpus. 
Since our method may involve multiple queries for one example, the number of its document candidates per example is from 50 to 100 (See Appendix~\ref{appendix:implementation_details}). For comprehensive comparison, we evaluate the re-ranking baselines with both 50 and 100 document candidates.
For the overall implementation details, please refer to Appendix~\ref{appendix:implementation_details}. 
\subsection{Main Results}
We show the results in Table~\ref{tab:main_result},
where we adopt the classification mode in retrieval verification for stable performance since the performance of score-and-filter mode varies with different thresholds.
We see that LLatrieval significantly outperforms baselines on all three datasets' correctness and verifiability, which shows its best overall retrieval performance. Specifically, LLatrieval outperforms the used retriever by 3.4 and 5.9 points on correctness and citation-F1, respectively, which straightly demonstrates its effectiveness. 
Notably, our method shows consistent improvements over other LLM-augmented retrieval methods, which only show comparable performance with traditional rerankers, and these indicate that our method can more effectively harness the LLM's abilities for augmenting retrieval. 
Meanwhile, various retrievers consistently show worse performance than LLM-augmented methods, and this demonstrates that the traditional retrievers become the bottleneck for verifiable generation and overshadow the LLM's strong abilities.
Additionally, our method demonstrates significant improvements over traditional neural rerankers, which indicates that the bottleneck effect of retriever can not be fully mitigated by traditional rerankers.
\subsection{Analyses}
In this section, we conduct analyses of LLatrieval, where we adopt classification-based retrieval verification and ``passage'' style missing-info query, and report Citation-F1 for verifiability for simplicity.
\paragraph{Ablation Study}
We evaluate the effectiveness of each component on ASQA and QAMPARI.
We show the performance changes after adding each component one-by-one, from the original retriever to the second iteration, in Table~\ref{tab:ablation_study}.
First we conduct progressive selection on the original question's document candidates and find it leads to performance improvements, which directly indicates that it can help the LLM find better relevant documents from the candidates. Then we verify whether each question's current documents can support answering it.
Based on the verification results, we divide them into two groups and separately evaluate their generation quality. We observe that on questions whose documents fail the verification, the LLM shows significantly worse generation quality in correctness and verifiability. And after retrieval update, the updated retrieval result leads to better generation quality.
These demonstrate that the retrieval verification and update can identify and improve low-quality retrieval results respectively, and thus lead to better verifiable generation.
In general, these results show that each component of LLatrieval contributes critically to the overall performance.

\begin{table}[]
\centering
\setlength\tabcolsep{1.8pt}
\small
\begin{tabular}{@{}lcccccc@{}}
\toprule
                        & \multicolumn{3}{c}{\textbf{ASQA}}             & \multicolumn{3}{c}{\textbf{QAMPARI}}        \\ \cmidrule(l){2-7} 
                        & \textbf{Num} & \textbf{EM-R} & \textbf{Cite} & \textbf{Num} & \textbf{F1} & \textbf{Cite} \\ \midrule
BGE-large               & 948          & 55.9          & 57.5           & 1000         & 17.3        & 24.0           \\
+ Progressive Selection & 948          & 56.8          & 60.2           & 1000         & 23.0        & 32.3           \\
\ \ \ \ + Verification {\color{green}\ding{51}}          & 779          & 60.3          & 64.5           & 827          & 25.8        & 36.6           \\
\ \ \ \ + Verification {\color{red}\ding{55}}          & 169          & 41.0          & 40.0           & 173          & 10.1        & 9.1            \\
\ \ \ \ \ \ \ \ + Update {\color{green}\rotatebox{90}{\ding{217}}}                & 169          & 44.3          & 44.3           & 173          & 17.7        & 21.1           \\ \bottomrule
\end{tabular}
\caption{The effect of each component. ``Verification {\color{green}\ding{51}} / {\color{red}\ding{55}} '' represent the question whose documents pass / fail the retrieval verification, respectively. ``Update {\color{green}\rotatebox{90}{\ding{217}}}'' represents the questions paired with updated retrieval result. ``Num'' is the corresponding example quantity.
}
\label{tab:ablation_study}
\end{table}

\paragraph{Performance over Different Thresholds} 

\begin{figure}[t]
    \centering
        \includegraphics[width=0.17\textwidth]{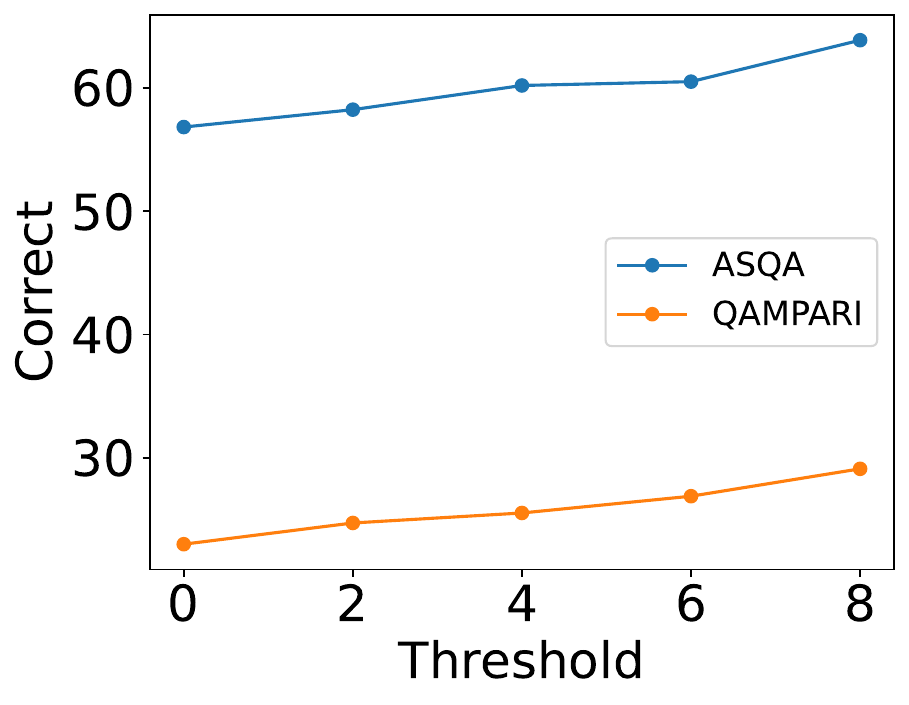}
        \includegraphics[width=0.17\textwidth]{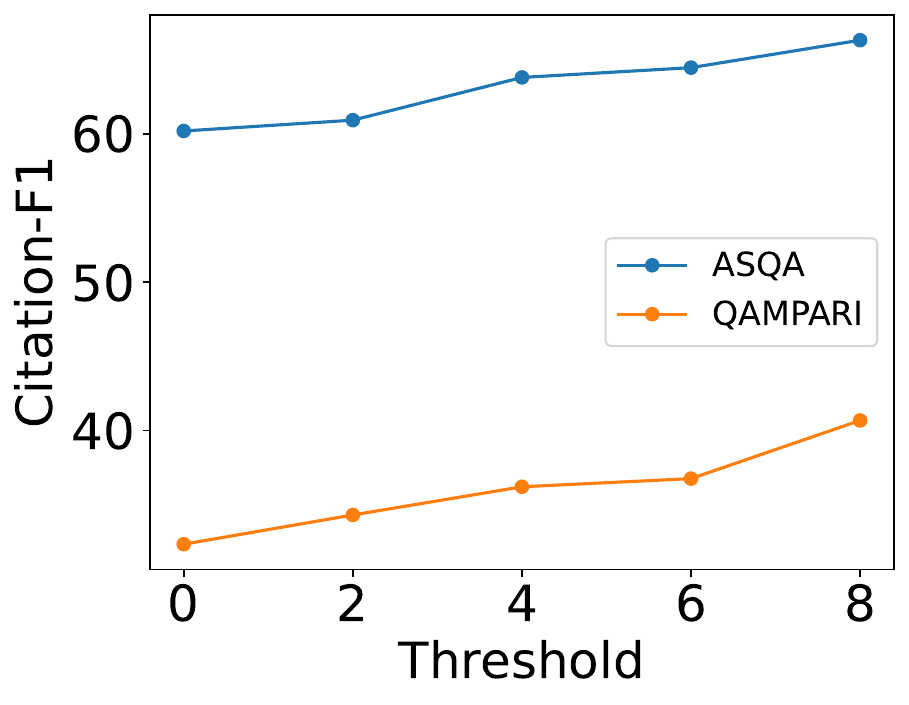}
    \caption{The generation quality of filtered examples over different thresholds.}
    \label{fig:filterd_performance_threshold}
\end{figure}

\begin{figure}[t]
    \centering
    \subfigure[ASQA-Correct]{
        \includegraphics[width=0.2\textwidth]{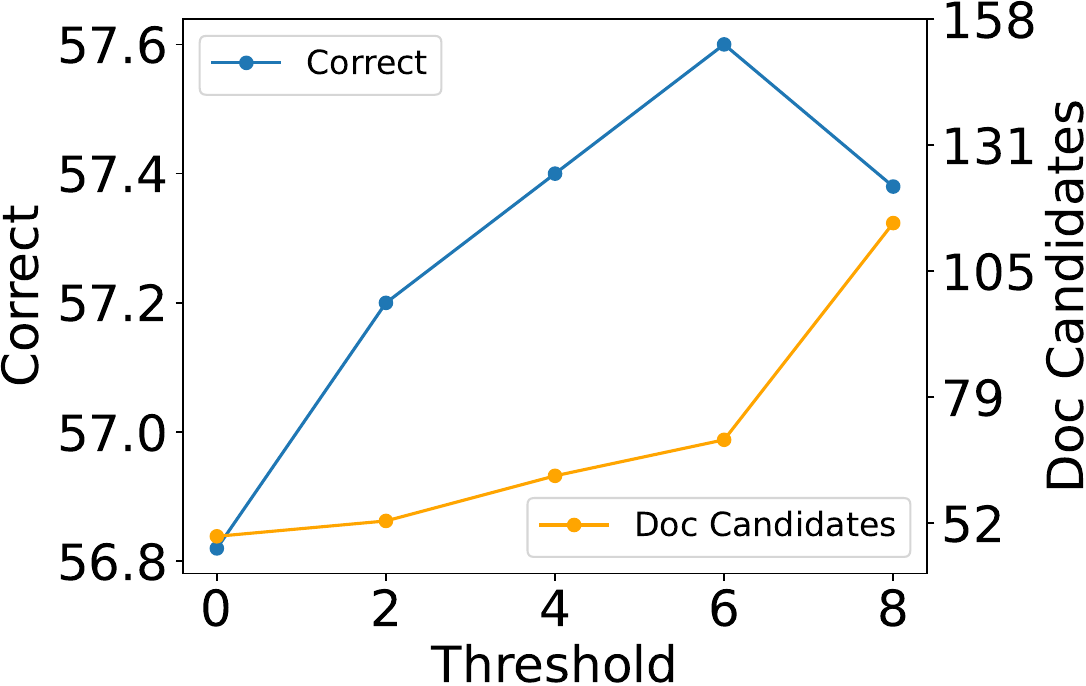}
    }
    \subfigure[ASQA-Citation]{
        \includegraphics[width=0.2\textwidth]{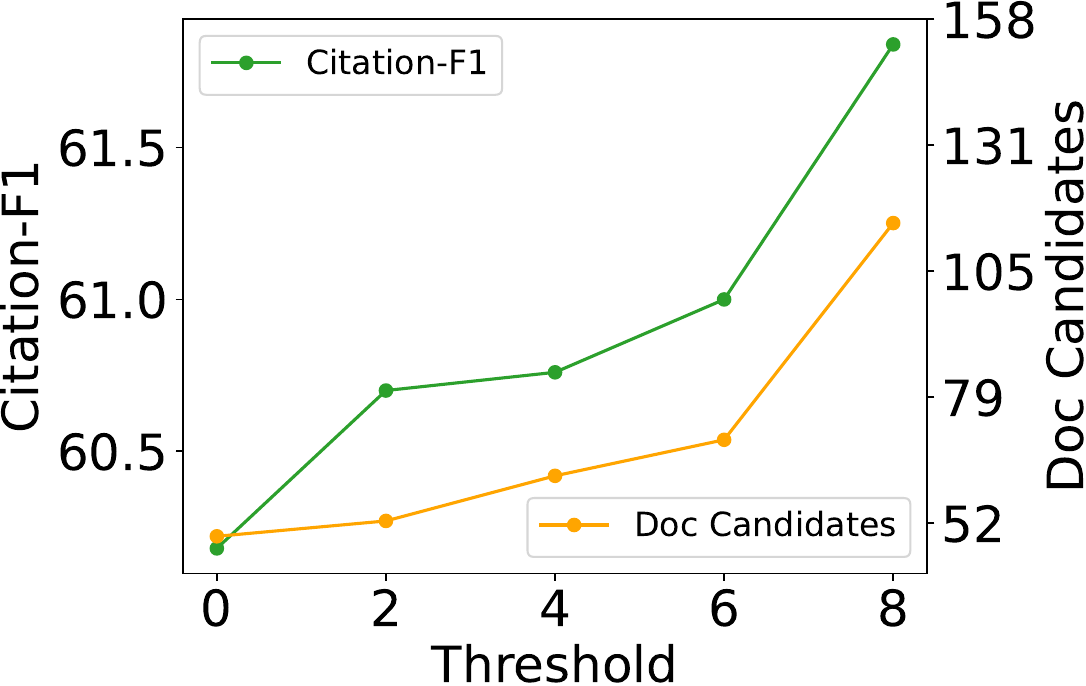}
    }
    
    \subfigure[QAMPARI-Correct]{
        \includegraphics[width=0.2\textwidth]{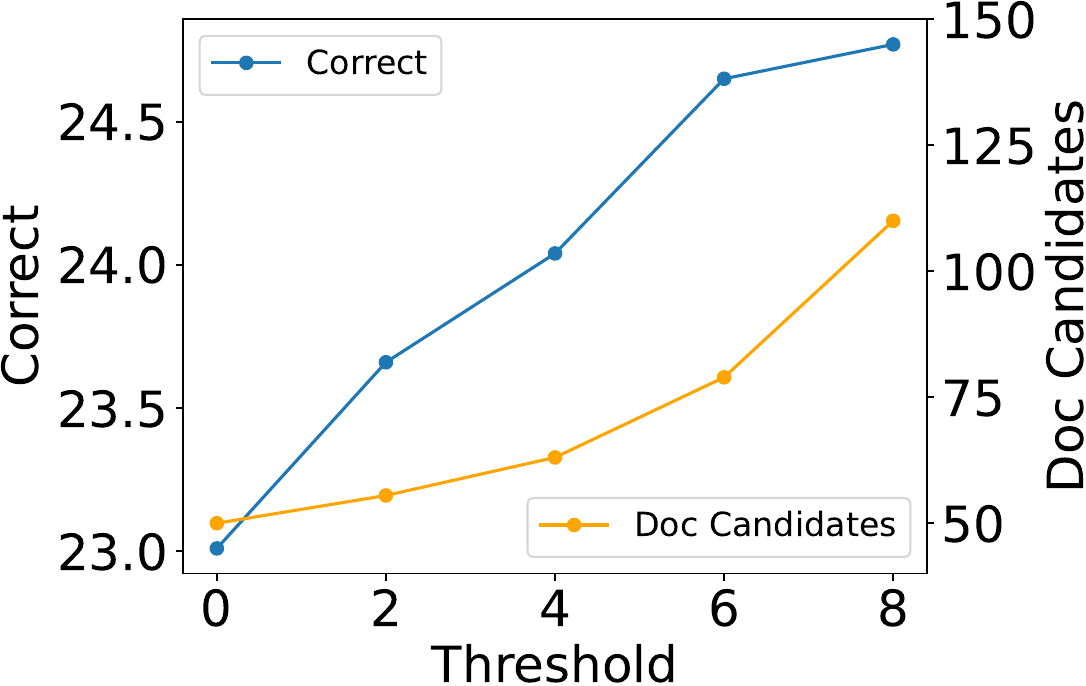}
    }
    \subfigure[QAMPARI-Citation]{
        \includegraphics[width=0.2\textwidth]{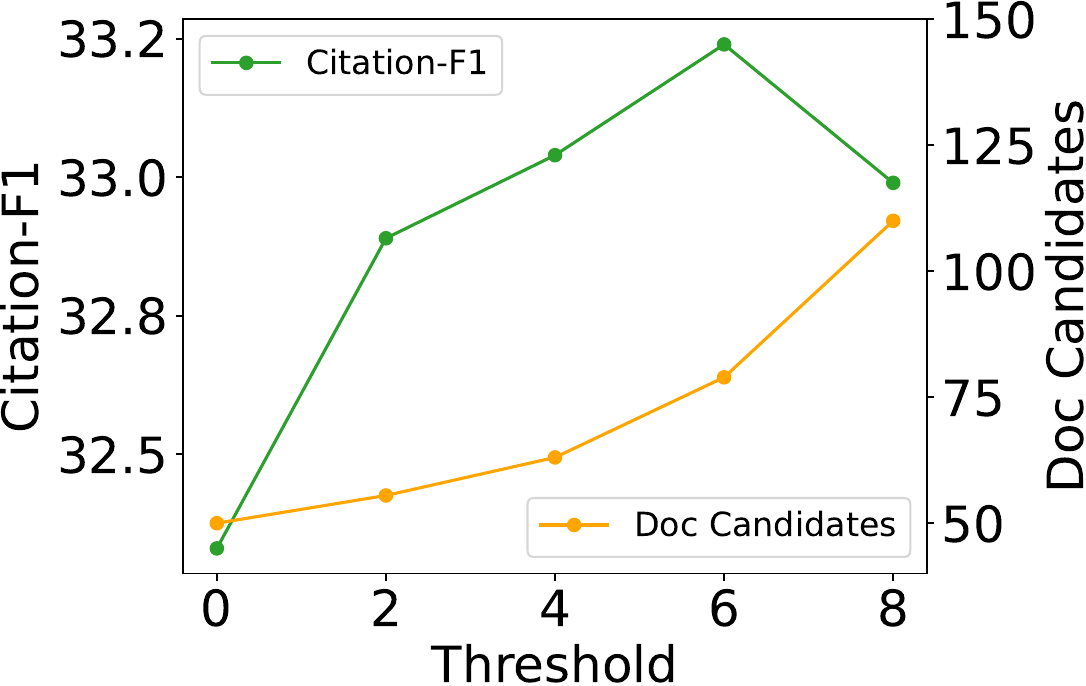}
    }
    \caption{The performance over different thresholds.}
    \label{fig:overall_performance_threshold}
\end{figure}

We evaluate the score-and-filter mode of retrieval verification on ASQA and QAMPARI. We first plot the generation quality on examples (question \& docs pairs) filtered by different thresholds in the first iteration, shown in Figure~\ref{fig:filterd_performance_threshold}. 
We see that the examples filtered by the higher threshold have better generation quality, which shows that our verification score can effectively evaluate the retrieval result. Then we plot the performance and the LLM cost of our method over different thresholds in Figure~\ref{fig:overall_performance_threshold}, where we use each example's overall document candidates quantity to measure the LLM cost since their tokens predominate in the LLM's input.
We see that increasing the threshold leads to better performance with higher LLM cost trade-off. This shows that our method's speed, i.e., average iterations, and LLM cost can be flexibly adjusted to adapt to varying demands.
Meanwhile, compared with reranking baselines, which do not show significant improvements with increasing document candidates (see Table~\ref{tab:main_result}), 
we can effectively scale our method's performance by stricter threshold and more document candidates from missing-info query. 
Additionally, we observe that a too-strict threshold may not necessarily improve performance, e.g., ``8'' for ASQA's correctness. We observe that its failure is because there is no relevant information in the document pool or the LLM fails to follow the instruction of missing-info query, and we regard the research of these two aspects as future work.

\paragraph{Performance during Iteration}
\begin{figure}[h]
    \centering
    \subfigure[ASQA-Correct]{
        \includegraphics[width=0.22\textwidth]{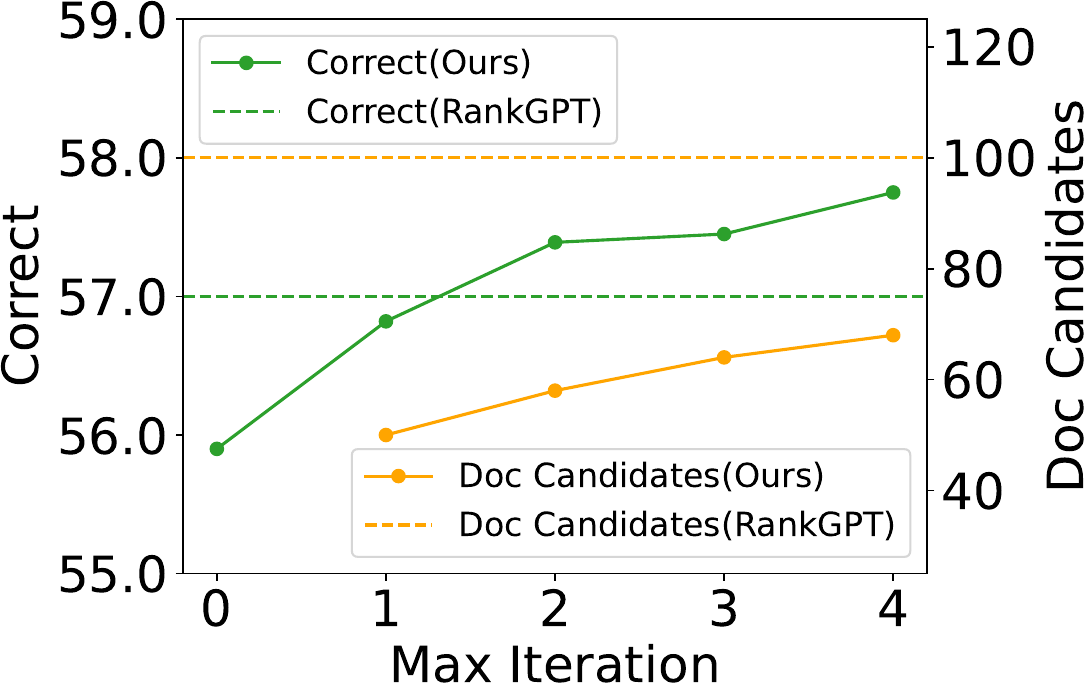}
    }
    \subfigure[ASQA-Citation]{
        \includegraphics[width=0.22\textwidth]{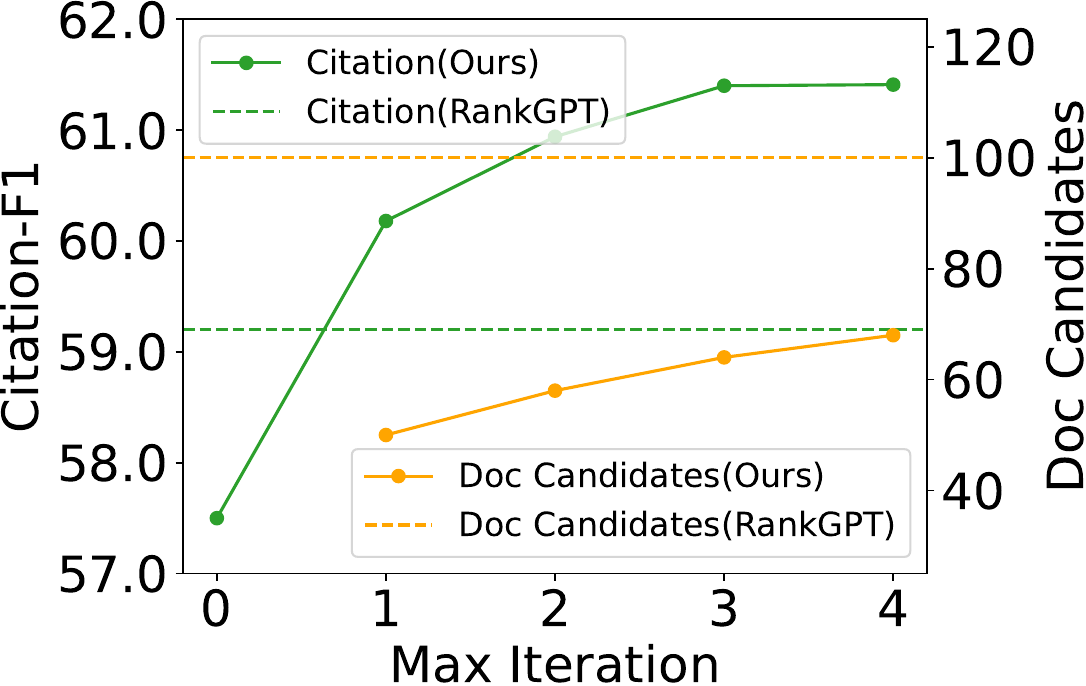}
    }
    \subfigure[QAMPARI-Correct]{
        \includegraphics[width=0.22\textwidth]{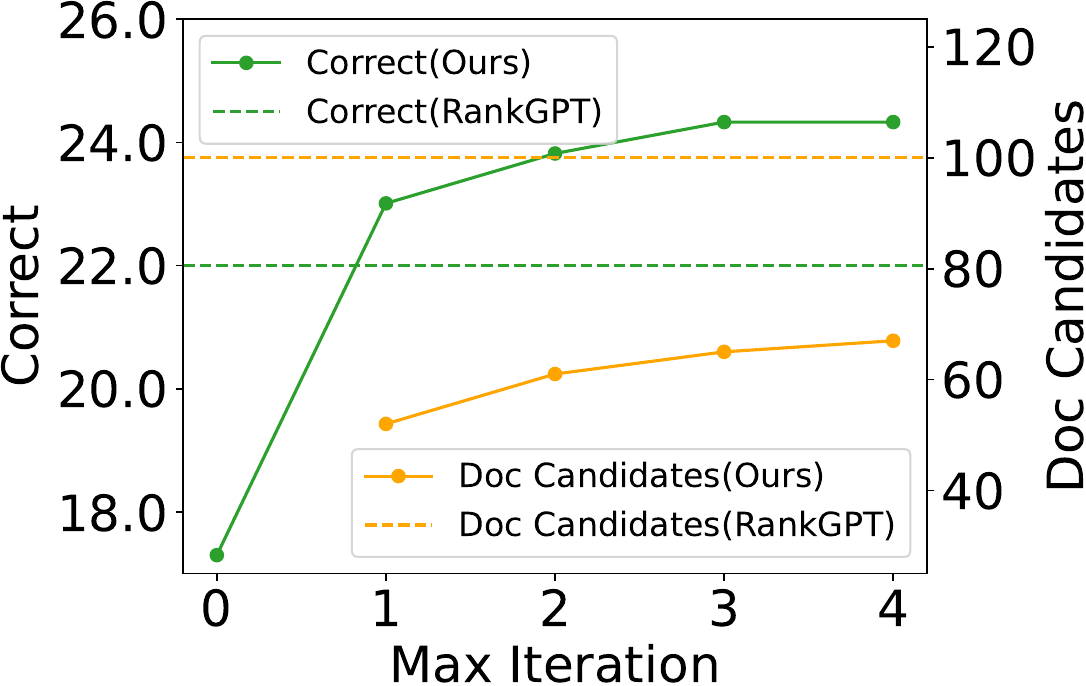}
    }
    \subfigure[QAMPARI-Citation]{
        \includegraphics[width=0.22\textwidth]{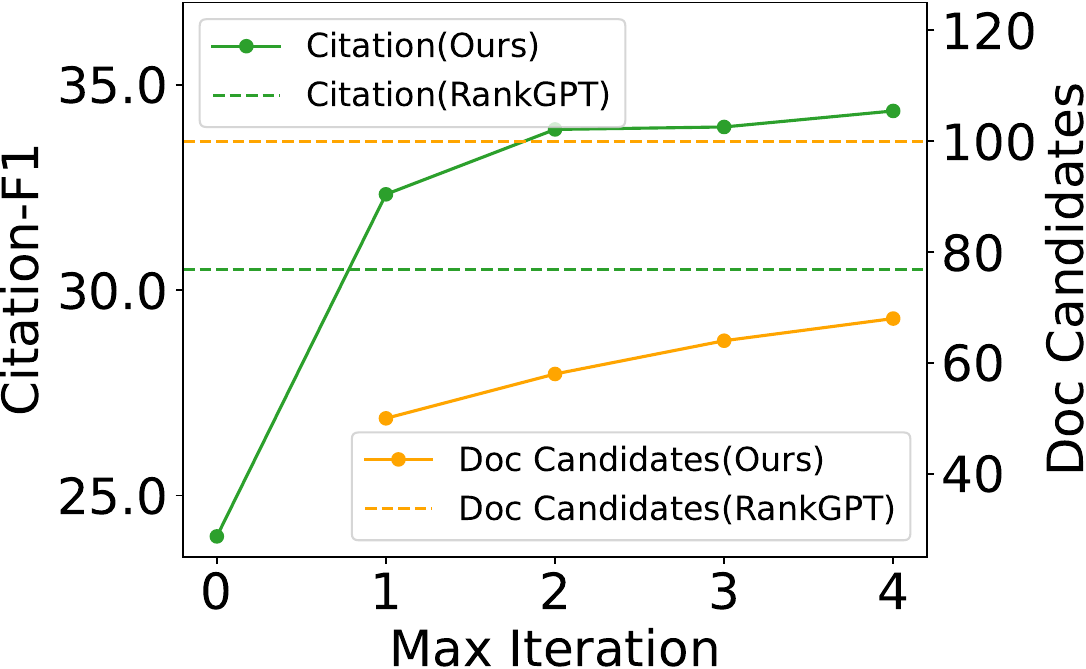}
    }
    \caption{The performance over different max iteration.}
    \label{fig:overall_performance_max_iteration}
\end{figure}
We observe the performance changes during verify-update iterations on ASQA and QAMPARI. Specifically, we plot the performance curve and each example's average document candidate quantity over different maximum iterations in Figure~\ref{fig:overall_performance_max_iteration}.
We see that the performance of LLatrieval gets improved with increasing max iterations, which shows that the document quality can be gradually improved during the multiple verify-update iterations.
Meanwhile, we observe that although the average document candidates slightly increase with increasing max iterations, our method can outperform RankGPT with fewer average document candidates, which shows LLatrieval's efficiency that it can dynamically finish when the retrieved documents can sufficiently support answering the question.

\begin{table}[]
\small
\centering
\setlength\tabcolsep{4pt}
\begin{tabular}{@{}lcccc@{}}
\toprule
\textbf{Dataset}   & \multicolumn{2}{c}{\textbf{QAMPARI}} & \multicolumn{2}{c}{\textbf{ELI5}} \\ \midrule
\textbf{Metric}    & \textbf{F1}   & \textbf{Cite}  & \textbf{Claim} & \textbf{Cite} \\ \midrule
Original Retriever & 17.3               & 24.0            & 15.2             & 66.7           \\ \midrule
GPT-3.5-Turbo-0301 & 24.3               & 34.4            & 16.5             & 71.3           \\
GPT-3.5-Turbo-0613 & 24.8               & 32.9            & 17.1             & 71.7           \\
GPT-3.5-Turbo-1106 & 23.9               & 33.6            & 17.3             & 71.7           \\
Llama-2-70B-Chat   & 21.0               & 28.9            & 15.8             & 61.0           \\
Xwin-LM-70B-V0.1   & 22.5               & 30.8            & 16.1             & 70.8           \\ 
Tulu-2-DPO-70B & 22.7 & 32.0 &16.9 & 71.8 \\
\bottomrule
\end{tabular}
\caption{The results of LLatrieval across various LLMs.}
\label{tab:various_llms}
\end{table}
\paragraph{The Comparison with Gold Feedback}
\begin{table}[]
\small
\centering
\begin{tabular}{@{}lccc@{}}
\toprule
                       & \multicolumn{3}{c}{\textbf{ASQA}}                          \\ \cmidrule(l){2-4} 
                       & \textbf{Doc Candidates} & \textbf{Correct} & \textbf{Cite} \\ \midrule
Our Method             & 68.9                    & 57.7             & 61.4          \\
Our Method (with gold)\hspace{-20pt} & 136.9                   & 58.1             & 60.8          \\ \midrule
                       & \multicolumn{3}{c}{\textbf{QAMPARI}}                       \\ \cmidrule(l){2-4} 
                       & \textbf{Doc Candidates} & \textbf{Correct} & \textbf{Cite} \\ \midrule
Our Method             & 64.5                    & 24.3             & 34.4          \\
Our Method (with gold)\hspace{-20pt} & 101.0                   & 25.2             & 33.6          \\ \bottomrule
\end{tabular}
\caption{The comparison with using gold answers for retrieval verification.}
\label{tab:external_feedback}
\end{table}
Recent work indicates that the feedback from the LLM can not help improve the prediction of itself~\citep{llm_cannot_self_correct_reasoning,gpt4_doesnot_know_its_wrong,can_llm_improve_by_self_critique}. 
Since we use the LLM to verify the retrieval result updated by itself, and such verification can be seen as a type of internal feedback, we compare it with the external feedback from gold answers on ASQA and QAMPARI. Specifically, for external feedback, we take whether gold answers fully appear in the retrieved result as verification criteria. We show the results in Table~\ref{tab:external_feedback}.
We see that both internal and external feedbacks achieve comparable performance with each other, when the internal feedback leads to less LLM cost.
These show that the LLM is capable of evaluating the retrieval result and has the potential to provide competitive feedback for retrieval as gold answers. 

\paragraph{The Generality across Various LLMs}
We analyze the performance of LLatrieval over various closed-source and open-source LLMs~\citep{llama2,llm_xwin,llm_tulu}
, shown in Table~\ref{tab:various_llms}. In general, our method consistently leads to significant improvements over the original retriever, which shows its robustness and generality.
Meanwhile, we find that our method with stronger LLM (from Llama2-70B to GPT-3.5) can lead to better performance, which indicates the potential to scale retrieval performance by scaling LLM and our method's utility in the future where more powerful LLMs are built, while dense retrievers have not been proved to scale well to the size of LLMs.

\section{Related Work}

\paragraph{Verifiable Generation}

Verifiable generation, which aims to generate content with supporting documents, has drawn increasing attention in recent years. The previous work mainly focuses on two aspects:
1) \textbf{Modeling:} \citet{web_gpt}, \citet{gopher_cite}, \citet{web_glm} and \citet{web_cpm} explore training the specialized LLM to browse web pages and answer long-form questions with supporting evidence. \citet{rarr} propose research-and-revision to retrieve supporting evidence for LLM's output and fix unsupported content. 
\citet{verifiable_generation_kg} explore verifiable generation on knowledge graph.
2) \textbf{Verifiability Evaluation:} \citet{attributed_lm_ais} define \textit{Attributable to Identified Sources} (AIS), a framework of human evaluation that considers whether the answer can be supported by the evidence. 
\citet{rarr} propose auto-AIS, which utilizes a strong NLI model~\citep{NLI_true} to approximate human AIS judgments. \citet{evaluating_verifiability_generative_search_engine} propose citation-recall and citation-precision to evaluate the verifiability of prevailing generative search engines, e.g., New Bing.
\citet{alce} propose a framework that automates the evaluation of recall and precision for the citation in the LLM's output.
Since retrieval is important for finding supporting evidence, we propose LLatrieval to iteratively update and verify the retrieval result, and thus fully harness the LLM for better verifiable generation.
\paragraph{LLM-based Retrieval Augmentation}
The existing methods of LLM-augmented Retrieval can be divided into three categories: 1) 
\textbf{Query Rewriting:}
\citet{hyde,query2doc} propose to use the LLM to generate a pseudo passage of the question as retrieval query, which neighbors relevant documents and helps retrieve them.
\citet{lamer} use the retrieved passage to help the LLM generate better pseudo passage as retrieval query.
\citet{retrieval_generation_synergy} propose to iteratively improve the retrieval query by  synergizing retrieval and generation.
\citet{query_rewriting_for_rag} train specialized query rewriters for downstream RAG tasks.
2) \textbf{Re-ranking:}
\citet{rank_gpt} propose RankGPT, which uses ChatGPT
to generate the permutation of documents for reranking. 
\citet{permutation_self_consistency} and \citet{pairwise_rank} use multiple-sampling and pair-wise ranking to mitigate the positional bias of RankGPT, respectively.
3) \textbf{Data Augmentation:}
\citet{inpars}, \citet{promptagator} and \citet{inparsv2} use the LLM to generate relevant queries for existing documents for training the retriever. \citet{art_retrieval} use the LLM to provide soft labels of unpaired question\&documents for the retriever's training.
Although these methods use the LLM to enhance the retrieval from various aspects, the LLM can not fully provide feedback to the retrieval result and the overall performance is highly limited by the retriever. For example, the query rewriting methods take the pseudo passage as query in a heuristic manner, which relies on the generated passage quality and can not ensure the relevance of retrieved documents. 
In LLatrieval, the LLM can iteratively provide feedback to the retriever until the documents sufficiently support verifiable generation, which can fully harness the LLM's ability to facilitate the retrieval stage.

\section{Conclusion}
We propose LLatrieval to retrieve supporting documents for verifiable generation, which lets the LLM iteratively refine the retrieval result until the LLM verifies the retrieved documents can support answering the given question. 
In this manner, LLatrieval can fully harness the LLM's ability and facilitate the retrieval result to help generate both correct and verifiable answers.
Experimental results show that our method outperforms extensive baselines and achieves new state-of-the-art results.
\section*{Limitations}
\begin{itemize}
    \item Although LLatrieval can dynamically adjust the running time according to the retrieval demand and shows better performance-efficiency trade-off than RankGPT~\citep{rank_gpt},
    it relies on real-time LLM inference and thus may not be suitable for tasks that require low latency or high throughput. However, over the years we have seen the model acceleration mechanisms advance and hardware performance increase, which can help improve the efficiency of LLM inference.
    \item Since existing works have shown LLMs exhibit various types of biases~\citep{llm_bias}, LLatrieval based on LLMs, may bias the final retrieval result. We are optimistic that this problem can be mitigated as LLatrieval is based on ChatGPT, and OpenAI has made substantial efforts to reduce its toxicity and bias~\citep{instruct_gpt}. Additionally, we can use more elaborate prompts on LLMs to further reduce the bias in the process of LLatrieval.

\end{itemize}

\section*{Acknowledgments}
This work was supported by the National Key Research and Development Program of China (No.2022CSJGG0801). The computations in this research were performed using the CFFF platform of Fudan University.

\bibliography{anthology,custom}
\clearpage
\appendix
\section{Implementation Details}
\label{appendix:implementation_details}
\subsection{Method Details}
We use the public OpenAI language model API of ``gpt-3.5-turbo-0301'' for retrieval and generation, unless otherwise specified. 
We set the temperature as 0 to mitigate the random fluctuation. 
We use window size=20 and set the number of document candidates per query as 50. 
We set the number of supporting documents and the maximum of iteration as 5 and 4. For the document pool, we follow \citet{alce} to use the Wikipedia for ASQA and QAMPARI, and Sphere~\citep{spere}, a filtered version of a filtered version of Common Crawl\footnote{\href{https://commoncrawl.org}{https://commoncrawl.org.}} for ELI5, respectively.
For ASQA and QAMPARI, we use the recently proposed dense embedder, BGE-large~\citep{bge_embedding}, as the retriever. For ELI5, we follow \citet{alce} to use BM25~\citep{bm25} since dense retrievers are costly and slow for large-scale web corpus. 

In progressive selection, we put the current documents before the new list of candidates in the LLM's input. Inspired by \citet{alce}, we use the document summary generated by LLM in progressive selection and thus we can significantly reduce the overall input length when the LLM select documents, which alleviates the challenges of LLMs finding relevant content in long context~\citep{lost_in_middle}. For ELI5, we use the HYDE~\citep{hyde} to generate the first iteration's query to get the better document candidates.
\paragraph{Compared Method Details}
For HYDE, LAMER and RankGPT~\citep{rank_gpt}, we use its original instructions and the LLM used in LLatrieval for fair comparison. For DPR, we follow \citet{alce} to use its original checkpoints on NQ~\citep{nq_dataset}.
For Contriever, we use its unsupervised version~\footnote{\href{https://huggingface.co/facebook/contriever}{https://huggingface.co/facebook/contriever}} and supervised version\footnote{\href{https://huggingface.co/facebook/contriever-msmarco}{https://huggingface.co/facebook/contriever-msmarco}} fine-tuned on MS MARCO. For GTR~\citep{gtr_embedding}, we follow \citet{alce} to use its T5-XXL verison\footnote{\href{https://huggingface.co/sentence-transformers/gtr-t5-xxl}{https://huggingface.co/sentence-transformers/gtr-t5-xxl}}. For instructor~\citep{instructor}, we use its 
large\footnote{\href{https://huggingface.co/hkunlp/instructor-large}{https://huggingface.co/hkunlp/instructor-large}} version. For BGE-Embedding, we use its 
large\footnote{\href{https://huggingface.co/BAAI/bge-large-en-v1.5}{https://huggingface.co/BAAI/bge-large-en-v1.5}} version.
Following~\citet{rank_gpt}, we use the 340M version of monoBERT\footnote{\href{https://huggingface.co/castorini/monobert-large-msmarco}{https://huggingface.co/castorini/monobert-large-msmarco}}, and 
3B\footnote{\href{https://huggingface.co/castorini/monot5-3b-msmarco-10k}{https://huggingface.co/castorini/
monot5-3b-msmarco-10k}} verison of monoT5.
For BGE-Reranker, we use its 
large~\footnote{\href{https://huggingface.co/BAAI/bge-reranker-large}{https://huggingface.co/BAAI/bge-reranker-large}} version for comparison.
\subsection{Instructions}
We show the overall instructions in Table~\ref{tab:instruction_retrieval_verification_classification}, \ref{tab:instruction_retrieval_verification_score}, 
\ref{tab:instruction_progressive_selection}, 
\ref{tab:instruction_missing_info_query_passage}, \ref{tab:instruction_missing_info_query_question}.
\subsection{Average Document Candidates}
In Table~\ref{tab:iteration_details_passage_style} and \ref{tab:iteration_details_question_style}, we show the average document candidate quantity and the number of examples that pass the retrieval verification in each iteration, for the result in Table~\ref{tab:main_result}.
\begin{table}[]
\small
\centering
\begin{tabular}{@{}lccc@{}}
\toprule
                    & \textbf{ASQA} & \textbf{QAMPARI} & \textbf{ELI5} \\ \midrule
Total Eexample      & 948           & 1000             & 1000          \\ \midrule
Document Candidates & 68.9          & 67.5             & 74.5 \\
\midrule
\multicolumn{4}{c}{\textbf{The number of examples in each iteration}}  \\ \midrule

Iteration 1         & 948           & 1000             & 1000          \\
Iteration 2         & 169           & 173              & 234           \\
Iteration 3         & 105           & 70              & 144           \\
Iteration 4         & 84            & 46               & 111           \\        \bottomrule
\end{tabular}
\caption{The iteration details and average document candidate quantity for our method (passage-style missing-info query).}
\label{tab:iteration_details_passage_style}
\begin{tabular}{@{}lccc@{}}
\toprule
                    & \textbf{ASQA} & \textbf{QAMPARI} & \textbf{ELI5} \\ \midrule
Total Eexample      & 948           & 1000             & 1000          \\ \midrule
Document Candidates & 69.2          & 66.9             & 76.0 \\
\midrule
\multicolumn{4}{c}{\textbf{The number of examples in each iteration}}  \\ \midrule

Iteration 1         & 948           & 1000             & 1000          \\
Iteration 2         & 169           & 173              & 234           \\
Iteration 3         & 112           & 91              & 159           \\
Iteration 4         & 83            & 73               & 128           \\        \bottomrule
\end{tabular}
\caption{The iteration detials and average document candidate quantity for our method (question-style missing-info query).}
\label{tab:iteration_details_question_style}
\end{table}

\begin{table}[]
\small
\centering
\setlength\tabcolsep{1.8pt}
\begin{tabular}{@{}lcccc@{}}
\toprule
\textbf{Dataset}     & \multicolumn{2}{c}{\textbf{ASQA}}        & \multicolumn{2}{c}{\textbf{QAMPARI}}     \\ \midrule
\textbf{Correctness} & \textbf{Retrieval} & \textbf{Generation} & \textbf{Retrieval} & \textbf{Generation} \\ \midrule
BM25                 & 47.7               & 51.7                & 29.7               & 17.5                \\
Instructor-large\hspace{-7pt}     & 57.0               & 55.6                & 27.8               & 17.2                \\
BGE-E-large\hspace{-3pt}          & 60.6               & 55.9                & 29.1               & 17.3                \\
monoBERT             & 60.8               & 53.8                & 35.7               & 20.9                \\
BGE-R-large\hspace{-3pt}          & 62.4               & 55.3                & 35.8               & 20.7                \\
RankGPT              & 62.1               & 56.4                & 34.4               & 20.7                \\
LLatrieval           & \textbf{62.9}               & \textbf{57.7}                & \textbf{38.3}               & \textbf{24.3}                \\ \bottomrule
\end{tabular}
\caption{The correlation between retrieval quality and generation quality.}
\label{tab_corrleation_retrieval_and_generation}
\end{table}

\begin{table*}[]
\small
\centering
\begin{tabular}{@{}lcccccccc@{}}
\toprule
\textbf{Dataset}                                                                      & \multicolumn{4}{c}{\textbf{ASQA}}                             & \multicolumn{4}{c}{\textbf{QAMPARI}}                          \\ \midrule
\multirow{2}{*}{\textbf{\begin{tabular}[c]{@{}l@{}}Evaluation\\ Metric\end{tabular}}} & \textbf{Correct} & \multicolumn{3}{c}{\textbf{Citation}}      & \textbf{Correct} & \multicolumn{3}{c}{\textbf{Citation}}      \\ \cmidrule(l){2-9} 
                                                                                      & \textbf{EM-R}    & \textbf{Rec} & \textbf{Prec} & \textbf{F1} & \textbf{F1}      & \textbf{Rec} & \textbf{Prec} & \textbf{F1} \\ \midrule
Original Retriever                                                                    & 55.9             & 56.4         & 58.6          & 57.5        & 17.3             & 22.7         & 25.5          & 24.0        \\ \midrule
Window Size=15                                                                        & 56.9             & 60.2         & 59.6          & 59.9        & 24.3             & 31.6         & 35.7          & 33.5        \\
Window Size=20                                                                        & 57.7             & 60.9         & 61.3          & 61.1        & 24.3             & 32.4         & 36.6          & 34.4        \\
Window Size=25                                                                        & 57.2             & 59.6         & 60.1          & 59.8        & 24.0             & 32.5         & 37.1          & 34.6        \\
Window Size=30                                                                        & 57.4             & 60.3         & 60.6          & 60.4        & 24.7             & 32.4         & 36.3          & 34.2        \\ \midrule
GTR                                                                                   & 54.4             & 52.2         & 54.6          & 53.3        & 18.5             & 20.0         & 23.2          & 21.5        \\
+ LLatrieval                                                                          & 57.6             & 59.0         & 60.5          & 59.7        & 22.6             & 27.1         & 31.9          & 29.3        \\
Instructor-large                                                                      & 55.6             & 52.1         & 54.1          & 53.1        & 17.2             & 19.4         & 21.9          & 20.6        \\
+ LLatrieval                                                                          & 57.6             & 60.1         & 60.2          & 60.2        & 21.1             & 24.7         & 29.1          & 26.7        \\
BGE-E-Base                                                                            & 55.3             & 56.2         & 58.2          & 57.2        & 17.9             & 22.9         & 24.8          & 23.8        \\
+ LLatrieval                                                                          & 56.7             & 58.5         & 59.3          & 58.9        & 25.0             & 32.6         & 37.3          & 34.8        \\ \midrule
Original Retriever ($|D|=3$)                                                            & 55.5             & 52.4         & 57.0          & 54.6        & 15.1             & 21.3         & 23.9          & 22.5        \\
+ LLatrieval ($|D|=3$)                                                                    & 56.6             & 57.9         & 61.9          & 59.9        & 23.1             & 31.3         & 35.0          & 33.0        \\
Original Retriever ($|D|=7$)                                                            & 56.9             & 59.0         & 58.4          & 58.7        & 17.5             & 22.6         & 25.6          & 24.0        \\
+ LLatrieval ($|D|=7$)                                                                    & 57.1             & 61.9         & 61.0          & 61.5        & 24.6             & 31.2         & 35.7          & 33.3        \\ \bottomrule
\end{tabular}
\caption{The performance of LLatrieval over various hyper-parameters and retrievers.}
\label{table_method_sensitivity}
\end{table*}

\section{The Correlation between Retrieval Quality and Generation Quality}
\label{appendix_correlation_retrieval_generation}
In exploratory experiments, we observe the correctness of retrieval result and LLM's generation when using various retrieval methods.
Specifically, we conduct experiments on ASQA and QAMPARI, and evaluate the exact match recall of retrieval result and the corresponding LLM's output, as shown in Table~\ref{tab_corrleation_retrieval_and_generation}. 
We see that the higher retrieval quality leads to higher generation quality, which shows the positive correlation between retrieval quality and the generation quality.

\section{Performance across Various Hyper-Parameters}
\label{appendix_method_sensitivity}
We run LLatrieval with different hyper-parameters and retrievers, and show the results in Table~\ref{table_method_sensitivity}. We see that LLatrieval consistently outperforms the original retriever under the different window sizes and document quantities. Meanwhile, LLatrieval can lead to performance improvements based on various retrievers. These show the robustness and generality of LLatrieval.

\begin{table*}[h]
    \centering
    \small
    \begin{tabular}{p{14cm}}

\toprule
You are JudgeGPT as introduced below.\\
\\
\textbf{\# Role: JudgeGPT}\\
\\
\textbf{\#\# Profile}\\
- Language: English\\
- Description: You are JudgeGPT, capable of judging whether a specified number (k) of documents can maximally support giving a direct, accurate, clear and engaging answer, similar to the answer of the demonstration, closely related to the core of the user's specific question(s).\\
\\
\textbf{\#\#\# Demonstration}\\
\{Demo\}\\
\\
\textbf{\#\#\# Input}\\
- Question: The specific question(s).\\
- Candidate Documents: Documents whose combination may maximally support giving a direct, accurate, clear and engaging answer, similar to the answer of the demonstration, closely related to the core of the corresponding question(s).\\
\\
\textbf{\#\#\# Skill}\\
1. Analyzing the given question(s) and understanding the required information.\\
2. Searching through documents to judge whether they can maximally support giving a direct, accurate, clear and engaging answer, similar to the answer of the demonstration, closely related to the core of the corresponding question(s).\\
\\
\textbf{\#\#\# Output}\\
\\
- Judgment: "[YES]" if provided documents can maximally support giving a direct, accurate, clear, and engaging answer, similar to the answer of the demonstration, closely related to the core of the corresponding question(s), otherwise "[NO]".\\
\\
\textbf{\#\#\# Output Format}\\
Judgment: [YES] or [NO]\\
\\
\textbf{\#\#\# Output Example}\\
If provided documents can maximally support giving a direct, accurate, clear, and engaging answer, similar to the answer of the demonstration, closely related to the core of the corresponding question(s), the output should be as follows:
[YES]\\
\\
\textbf{\#\# Rules}\\
1. Don't break character.\\
2. When outputting final verdict, only providing "[YES]" or "[NO]".\\
3. Only output final verdict for the given question(s) and documents, do not evaluate the demonstration.\\
4. Strictly follow the specified output format. Do not answer the given question. Just conduct the specified judgment task.\\
\\
\textbf{\#\# Judgment Criteria (Very Important)}\\
1. Do not allow the length of the documents to influence your evaluation.\\
2. Be as objective as possible.\\
3. Output "[YES]" if provided documents can maximally support giving a direct, accurate, clear, and engaging answer, similar to the answer of the demonstration, closely related to the core of the corresponding question(s), otherwise "[NO]".\\
\\
\textbf{\#\# Workflow}\\
1. Read and understand the questions posed by the user.\\
2. Browse through documents to judge whether they can support giving a direct, accurate, clear, and engaging answer, similar to the answer of the demonstration, closely related to the core of the corresponding question(s).\\
3. Output your final verdict.\\
\\
\textbf{\#\# Reminder}\\
You will always remind yourself of the role settings.\\

        \bottomrule
    \end{tabular}
\caption{
    The instruction for retrieval verification based on classification.
    }
    \label{tab:instruction_retrieval_verification_classification}

\end{table*}

\begin{table*}[h]
    \centering
    \small
    \begin{tabular}{p{14cm}}

\toprule
You are ScoreGPT as introduced below.\\
\\
\textbf{\# Role: ScoreGPT}\\
\\
\textbf{\#\# Profile}\\
- Language: English\\
- Description: You are ScoreGPT, capable of scoring candidate documents based on their level of support for the corresponding question(s), with a rating range from 0 to 10.\\
\\
\textbf{\#\#\# Input}\\
- Question: The specific question(s).\\
- Candidate Documents: Documents whose combination may maximally support the corresponding question(s).\\
\\
\textbf{\#\#\# Skill}\\
1. Analyzing the given question(s) and understanding the required information.\\
2. Searching through documents to score them based on their level of support for the corresponding question(s), with a rating range from 0 to 10.\\

\textbf{\#\#\# Output}\\
- A score ranging from 0 to 10, where a higher score indicates greater support of the candidate documents for the corresponding question(s), and a lower score indicates lesser support.\\

\textbf{\#\#\# Output Format}\\
$\left[\text{SCORE}\right]$\\
\\
\textbf{\#\# Rules}\\
1. Don't break character.\\
2. When outputting final score, only providing "[SCORE]".\\
3. Strictly follow the specified output format. Do not answer the given question(s). Just conduct the specified scoring task.\\
\\
\textbf{\#\# Score Criteria (Very Important)}\\
1. Do not allow the length of the documents to influence your evaluation.\\
2. Be as objective as possible.\\
3. Output "[SCORE]" ranging from 0 to 10, where a higher score indicates greater support of the candidate documents for the corresponding question(s), and a lower score indicates lesser support.\\
\\
\textbf{\#\# Workflow}\\
1. Read and understand the question(s) posed by the user.\\
2. Browse through documents to score them based on their level of support for the corresponding question(s), with a rating range from 0 to 10.\\
3. Output your final score surrounded by square brackets.\\
\\
\textbf{\#\# Reminder}\\
You will always remind yourself of the role settings.\\

        \bottomrule
    \end{tabular}
\caption{
    The instruction for retrieval verification based on score-and-filter.
    }
    \label{tab:instruction_retrieval_verification_score}

\end{table*}

\begin{table*}[h]
    \centering
    \small
    \begin{tabular}{p{14cm}}

\toprule
You are DocSelectorGPT as introduced below.\\
\\
\textbf{\# Role: DocSelectorGPT}\\
\\
\textbf{\#\# Profile}\\
- Language: English\\
- Description: You are DocSelectorGPT, capable of selecting a specified number (k) of documents for answering the user's specific question(s). k is a value specified by the user.\\
\\
\textbf{\#\#\# Input}\\
- Question: The specific question(s)\\
- Candidate Documents: Documents contain supporting documents which can support answering the given questions. Candidate documents will have their own identifiers for FactRetrieverGPT to cite.\\
\\
\textbf{\#\#\# Skill}\\
1. Analyzing the given question(s) and understanding the required information.\\
2. Searching through candidate documents to select k supporting documents whose combination can maximally support giving a direct, accurate, clear and engaging answer to the question and make the answer and is closely related to the core of the question.\\
\\
\textbf{\#\#\# Output}\\
- Selected Documents: The identifiers of selected supporting documents whose combination can maximally support giving an accurate and engaging answer to the question and make the answer and is closely related to the core of the question.\\
\\
\textbf{\#\#\# Output Format}\\
Selected Documents: [document identifiers]

\textbf{\#\#\# Output Example}\\
If the selected documents are 2, 6 and 8, the output should be as follows:\\

Selected Documents: 2 6 8\\
\\
\textbf{\#\# Rules}\\
1. Don't break character.\\
2. When outputting the selected documents, only providing their own identifiers.\\
3. Strictly follow the specified output format. Do not answer the given question. Just conduct the specified retrieval task.\\
\\
\textbf{\#\# Selection Criteria (Very Important)}\\
1. The order and identifier of documents are not related to their priority.\\
2. Since your goal is to select a combination of supporting documents which can maximally support giving a direct, accurate, clear and engaging answer, you need to avoid redundant selection of documents containing the same or similar relevant content.\\
\\
\textbf{\#\# Workflow}\\
1. Read and understand the questions posed by the user.\\
2. Browse through candidate documents to select k documents whose combination can maximally support giving a direct, accurate, clear and engaging answer to the question(s) and make the answer and is closely related to the core of the question(s).\\
3. List all selected documents.\\
\\
\textbf{\#\# Reminder}\\
You will always remind yourself of the role settings.\\

        \bottomrule
    \end{tabular}
\caption{
    The instruction for LLM to select documents in progressive selection.
    }
    \label{tab:instruction_progressive_selection}

\end{table*}

\begin{table*}[h]
    \centering
    \small
    \begin{tabular}{p{14cm}}

\toprule

You are a helpful assistant as introduced below.\\
\\
\textbf{\#\# Profile}\\
- Language: English\\
- Description: You are a helpful assistant, capable of identifying missing content that answers the given question(s) but does not exist in the given possible answering passages and then using your own knowledge to genereate correct answering passages using missing content you identify.\\
\\
\textbf{\#\#\# Input}\\
- Question: The specific question(s).\\
- Answering Passages: Possible answering passages.\\
\\
\textbf{\#\#\# Output}\\
- Correct answering passages generated using missing content you identify based on your own knowledge.\\
\\
\textbf{\#\# Rules}\\
1. Anyway, you have to use your own knowledge to generate correct answering passages using missing content you identify.\\
2. Only generate the required correct answering passages. Do not output anything else.\\
3. Directly use your own knowledge to generate correct answering passages if you think the given possible answering passages do not answer to the given question(s). \\
4. Do not output the given question(s) and possible answering passages.\\
5. Do not output your analysis statement.\\
\\
\textbf{\#\# Workflow}\\
1. Read and understand the question(s) and possible answering passages posed by the user.\\
2. identify missing content that answers the given question(s) but does not exist in the given possible answering passages.\\
3. Directly use your own knowledge to generate correct answering passages if you think the given possible answering passages do not answer to the given question(s). Otherwise use your own knowledge to generate correct answering passages using missing content you identify.\\
\\
\textbf{\#\# Reminder}\\
You will always remind yourself of the role settings.\\
        
        \bottomrule
    \end{tabular}
\caption{
    The Instruction for Missing-info Query (the passage style).
    }
    \label{tab:instruction_missing_info_query_passage}

\end{table*}

\begin{table*}[h]
    \centering
    \small
    \begin{tabular}{p{14cm}}

\toprule

You are a helpful assistant as introduced below.\\
\\
\textbf{\#\# Profile}\\
- Language: English\\
- Description: You are a helpful assistant, capable of identifying missing content that answers the given question(s) but does not exist in the given possible answering passages and then using your own knowledge to genereate a new question based on the missing content you identify.\\
\\
\textbf{\#\#\# Input}\\
- Question: The specific question(s).\\
- Answering Passages: Possible answering passages.\\
\\
\textbf{\#\#\# Output}\\
- A new question generated using missing content you identify based on your own knowledge.\\

\textbf{\#\# Rules}\\
1. Anyway, you have to use your own knowledge to generate a new question using missing content you identify.\\
2. Only generate the required new question. Do not output anything else.\\
3. Do not output the given question(s) and possible answering passages.\\
4. Do not output your analysis statement.\\
\\
\textbf{\#\# Workflow}\\
1. Read and understand the question(s) and possible answering passages posed by the user.\\
2. Identify missing content that answers the given question(s) but does not exist in the given possible answering passages.\\
3. Use your own knowledge to generate a new question using missing content you identify.\\
\\
\textbf{\#\# Reminder}\\
You will always remind yourself of the role settings.\\
        
        \bottomrule
    \end{tabular}
\caption{
    The instruction for Missing-info Query (the question style).
    }
    \label{tab:instruction_missing_info_query_question}

\end{table*}

\end{document}